\documentclass[11pt]{article}
\usepackage[a4paper]{geometry}
\usepackage{clic2017}
\usepackage{times}
\usepackage{url}
\usepackage{latexsym}
\usepackage{xcolor}
\usepackage{graphicx}

\title{A Novel Integrated Industrial Approach with Cobots in the Age of Industry 4.0 through Conversational Interaction and Computer Vision}

 \author{Andrea Pazienza \and Nicola Macchiarulo \and Felice Vitulano  \and \\
         {\bf Alessio Germinario} \and 
         {\bf Antonio Fiorentini} \and {\bf Marco Cammisa} 
         \\ Innovation Lab, Exprivia S.p.A. \\  
         {\tt \{andrea.pazienza, nicola.macchiarulo, felice.vitulano,}\\
         {\tt alessio.germinario,
         antonio.fiorentini, marco.cammisa\}@exprivia.com } 
         \AND
         Leonardo Rigutini \and Ernesto Di Iorio \and Achille Globo \and Antonio Trevisi
         \\ QuestIT S.r.l. \\ 
         {\tt \{rigutini, diiorio, globo, trevisi\}@quest-it.com }
         }
 
\date{ }

\begin{document}
\maketitle
\begin{abstract}
  \textbf{English.} 
  From robots that replace workers to robots that serve as helpful colleagues, the field of robotic automation is experiencing a new trend that represents a huge challenge for component manufacturers.
  The contribution starts from an innovative vision that sees an ever closer collaboration between Cobot, able to do a specific physical job with precision, the AI world, able to analyze information and support the decision-making process, and the man able to have a strategic vision of the future.
\end{abstract}


\section{Introduction}
{\let\thefootnote\relax\footnote{{Copyright \textcopyright \  2019 for this paper by its authors. Use permitted under Creative Commons License Attribution 4.0 International (CC BY 4.0).}}}
In the last century, the manufacturing world has adopted solutions for the advanced automation of production systems. 
Today, thanks to the evolution and maturity of new technologies such as Artificial Intelligence (AI), Machine Learning (ML), new generation networks, and the growing adoption of the Internet of Things (IoT) approach, a new paradigm emerges, aiming at integrating the Cyber-Physical System (CPS) with business processes, thus opening the doors to the fourth industrial revolution (Industry 4.0) and that will allow us to join in the era driven by information and further handled with cognitive computing techniques~\cite{wenger2014artificial}.

Robots and humans have been co-workers for years, but rarely have we been truly working together. This may be about to change with the rise of Collaborative Robotics~\cite{colgate1996cobots}.
Collaborative Robots (better known as \textbf{Cobots}) are specifically designed for direct interaction with a human within a defined collaborative work-space, i.e., a safeguard space where the robot and a human can perform tasks simultaneously during an automatic operation. Then, human-robot collaboration fosters various levels of automation and human intervention. Tasks can be partially automated if a fully automated solution is not economical or too complex. 
%
Therefore, manufacturers may benefiting from the rising of AI-driven automation, and the progress of \textbf{Adaptable End Effectors}  devices, mounted at the end of Cobot's arms, may help to perform specific intelligent tasks~\cite{dubey2002finger}. 

The way in which Cobots and humans interact, exchanging and conveying information, is fundamental. The key role in this landscape would be addressed by \textbf{Conversational Interfaces}~\cite{zue2000conversational}, which exploit and take advantages from the recent achievements in the field of Natural Language Processing (NLP), to understand user need and generate the right answer or action.
%
In this scenario, \textbf{Computer Vision} also plays an important role in the process of creating collaborative environments between humans and robots. Systems of this type are already introduced into the industry to facilitate tasks of product quality control or component assembly inspection. By giving vision to a robot, it can make it able to understand the industrial environment that surrounds it and can improve the execution of tasks in support to other people.

Improving robots’ software with AI will be key to making robots more collaborative. The work starts from an innovative vision that beholds, in the future, an ever closer collaboration between Cobot, able to do a specific physical job with precision and without alienation, the AI world, able to analyze, process, and learn from information and support the decision-making process, and the employee able to have a strategic vision of the future. 
To validate its effectiveness, a collaborative environment between employee, Cobot and AI systems has been crafted to make possible the three subjects communicate in a simple way and without requiring the employee to have specific skills to interact with the Cobot and Enterprise Resource Planning (ERP) systems.

Our contribution is indeed placed in this scenario where the convergence of multiple technologies allows us to define a new approach related to the management of a core business process (e.g. shipments) which tends to ensure more and more flexibility of the process thanks to a simplification of human interaction with Cyber-Physical Systems, with a better coordination between the physical world (the packaging line), and that of IT processes (the ERP model). 
In the belief that the complexity of new industrial production systems requires interdisciplinary skills, our intents are to bring together knowledge from related disciplines such as computational linguistics, cognitive science, machine learning, computer vision, human-machine interaction, and collaborative robotics automation towards an integrated novel approach specifically designed for the smart management of a manufacturing process line by fostering and strengthening the synergy and the interaction between robot and human.

Our research is broadly situated in Human–Robot Collaboration (HRC), a promising robotics discipline focusing on enabling robots and humans to operate jointly to complete collaborative tasks. Recent works tried to figure out in which way Cobots may help humans in collaborative industrial tasks~\cite{el2019cobot} or in participatory design in fablabs~\cite{ionescua2019participatory}. An inital study centered cobots in advanced manufacturing systems~\cite{djuric2016cobot}.
No or litte work~\cite{ivorra2018multimodal} is done to endow Cobots with cognitive intelligence like conversational interaction and computer vision.

This paper is organized as follows. 
Section~\ref{sec:cob} introduces the functionalities and the architecture of our approach, focusing on the main four technological aspects: cobots, adaptable end effectors, conversational interfaces, and computer vision. 
Section~\ref{sec:scenarios} describes the possible scenarios of application specifically designed for our approach, such as Smart Manufacturing.
Finally, Section~\ref{sec:conclusion} discusses the proposed framework and concludes the paper, outlining future works.

\section{Architecture Proposal}\label{sec:cob}
\begin{figure*}[tb]
    \centering
    \includegraphics[scale=.6]{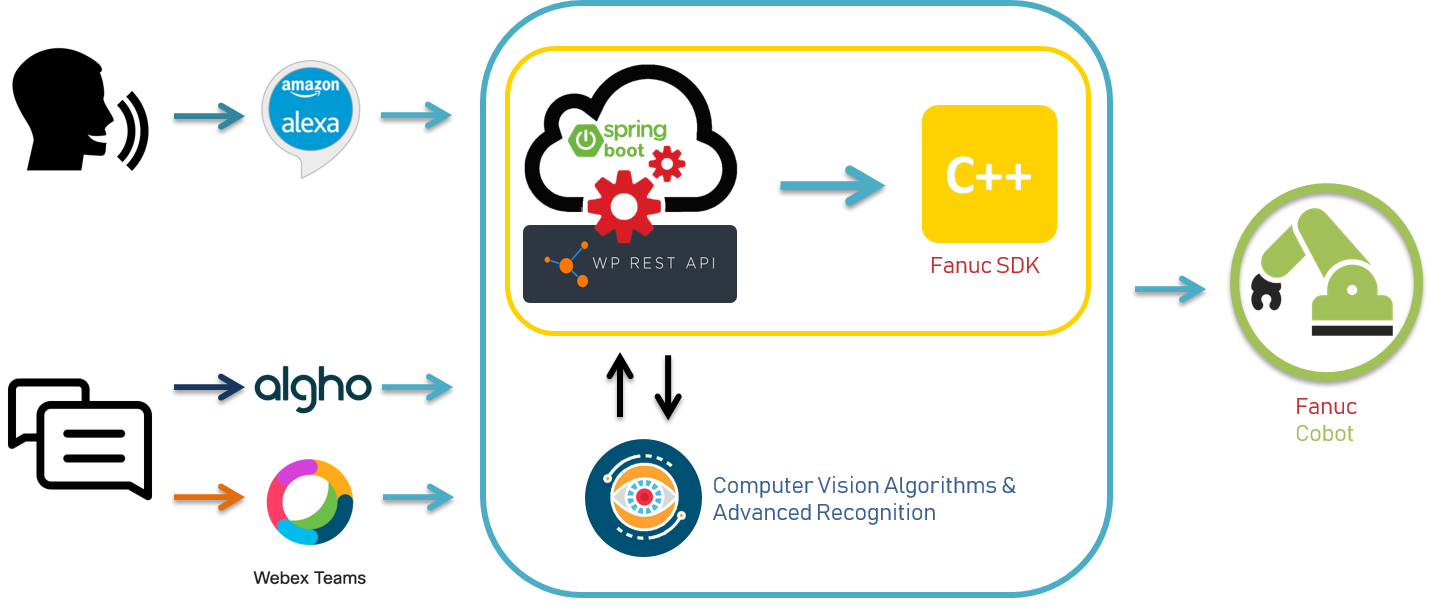}
    \caption{Framework Architecture including Conversational Interfaces, Computer Vision and Cobot}
    \label{fig:framework}
\end{figure*}

In this Section we introduce our main proposal taking into account all the requirements coming from different technologies. The leading idea is to develop and validate a general framework concerning an Intelligent Cyber-Physical System made up of four crucial components: (i) a Cobot, equipped with (ii) an adaptable end effector, which may change according to a specific scenario, and two major components coming from the AI world, i.e. (iii) a Computer Vision module to allow the cobot detecting an object, and (iv) one or more Conversational Interfaces to facilitate the human-machine interaction and keep the man in the loop. 
Figure~\ref{fig:framework} depicts the prototypical architecture of our framework proposal.

In order to integrate different technologies, from the high-level voice command to low-level execute command, we developed a web application, powered by \emph{Spring Boot} framework\footnote{https://spring.io/projects/spring-boot}, able to receive commands from user interfaces and transform them into machine commands.
We consider in this framework the possibility to give vocal commands to the Cobot. In this perspective, the mechanical arm is controlled through a series of connected conversational devices like chat-bots, powered by \emph{Cisco WebEx Teams}\footnote{https://www.webex.com/team-collaboration.html} and \emph{QuestIT Algho}\footnote{\label{algho}https://www.alghoncloud.com/it/}, and a virtual assistant such as \emph{Amazon Alexa}\footnote{https://developer.amazon.com/it/alexa}. In particular, Cisco WebEx Teams is an all-in-one solution for messaging, file sharing, white boarding, video meetings, and calling, while Amazon Alexa is a voice interaction device capable of a large set of human-interacting functions. 
The Cobot, through the use of a camera, is able to acquire images and process through the use of Computer Vision algorithms, recognizing exactly the object to be selected without knowing its position in advance.
Hence, the vocal commands sent via Alexa are managed by lambda functions using the \emph{AWS Lambda} service\footnote{https://aws.amazon.com/it/lambda/}, which is a serverless event-driven computing platform. It permits to execute code in response to particular events, automatically managing also the resources required by the programming code. Indeed, lambda's goal is to simplify the construction of on-demand applications that respond to events and new information.

Therefore, all commands are sent via HTTP calls to the web application using the Spring Boot framework, receiving also calls from the one or more chat-bots with which the user can interact.
Once a command has been received, the web application executes a C\# application, based on Fanuc SDK, that sends to the Cobot the request to execute a particular script written in Teach Pendant (TP) language.

\subsection{Cobot with Fanuc}
The right choice of a cobot comes with the fulfillment of various safety requirements, such as a collision stop protection, a function to restart them easily and quickly after a stop, and anti-trap features for additional protection.
For our purposes, we used a Cobot from Fanuc, in particular the \emph{CR-4iA} model\footnote{https://www.fanuc.eu/it/it/robot/robot-filter-page/robot-collaborativi/collaborative-cr4ia}. It is endowed with six axis in its arm, and its maximum payload is $4$ kg. Also, it handles lightweight tasks that are tedious, highly manual. 
Since it can take over these dull jobs, the operator hands are free to focus on more intelligent work or even more pressing matters. This cobot can also work side-by-side on tasks that are more complex, and require more interactive approaches. 
%

\subsection{Adaptable End Effector with Schunk}
This category includes grippers, which hold and manipulate objects, and end-of-arm tools (EOATs), which are complex systems of grippers designed to handle large or delicate components.
Handling tasks mainly include pick and place, sorting, packaging, and palletizing. 
As gripping tool we used the \emph{Schunk Co-act EGP-C} gripper\footnote{https://schunk.com/it\_it/co-act/pinza-co-act-egp-c/}. It is an Electric 2-finger parallel gripper certified for collaborative operation with actuation via 24 V and digital I/O.
It is used for gripping and moving small and medium-sized workpieces with flexible force in collaborative operation in the areas of assembly, electronics and machine tool loading.
We chose this model due to its certified and pre-assembled gripping unit with funcional safety, and its ``plug \& work'' mode with Fanuc cobots.

\subsection{Conversational Interfaces with Algho}
The achievements in the field of Artificial Intelligence (AI) in the recent years have led to the birth of a new paradigm of human-machine interaction: the conversational agents.
This new way of interacting with a computer is based on the use of natural language and is getting closer to the way humans communicate with each other.
Conversational agents take advantage of recent achievements in the field of Natural Language Processing (NLP-U) to understand user requests and behave accordingly, providing appropriate answers or performing required actions.
The design of an innovative Cobot cannot fail to consider the use of a such straightforward human-machine interaction.

The conversational functionalities for the Cobot described in this paper have been provided by using Algho\footnotemark{\ref{algho}}, a proprietary conversational-agent building tool developed by QuestIT\footnote{\label{qit}https://www.quest-it.com} and based on NLP and AI techniques.
In particular, Algho is a suite designed to facilitate the creation of personal conversational agents and the subsequent deploy on several proprietary channels. 
The user of Algho can create his own chat-bot simply by entering the personal knowledge base and the system, after a few minutes, is able to handle conversations about it.
The natural language understanding functionalities of Algho are based on a proprietary NLP Platform developed by QuestIT\footnotemark{\ref{qit}} consisting in more than $25$ layers of morphological, syntactic and semantic analysis based on Machine Learning (ML) and Artificial Intelligence techniques: tokenization, lemmatization, Part-Of-Sopeech (POS), Collocation Detection, Word Sense Disambiguation, Dependency Tree Parsing, Sentiment and Emotional Analysis, Intent Recognition, and many others.
The NLP Platform exploits the most recent techniques in the field of NLP and Machine Learning to enrich the input raw text with a set of high-level cognitive information~\cite{rigutini2018enhancingwsd,rigutini2018recursive}.
The Word Sense Disambiguation (WSD) layer is one of the main levels of the NLP Platform and it follows a Deep Neural Network approach based on RNN and word embedding. It provides state-of-the-art performances with regard to the disambiguation accuracy~\cite{rigutini2018enhancingwsd,rigutini2018recursive}.

The enriched text is subsequently exploited by the conversational engine to understand the user request, to identify the ``intent'' and to behave accordingly to the knowledge base provided by the creator of the conversational agent.
The intent of a request is defined as the hidden desire that underlies the user's request. 

During the construction of the conversational agent, the Algho suite allows the user to define specific objects called ``Conversational Form'' which can be used to collect structured information from the user. 
In particular, a ``conversational form'' consists in a typical form for collecting data which is linked to a set of intent defined in the knowledge base. 
During the conversation, when an input user request triggers an intent having a linked conversational form, the system:
    (i) tries to fill the form fields by extracting the information from the NLP analysis of the request (Auto-Form-Filling procedure); 
    and 
    (ii) proposes sequentially to the user the fields that have not been filled by the automatic procedure.
When an user input request trigger a conversational form, the returned NLP information are used to automatically fill the fields of the structured form without requesting further data from the user.
Furthermore, Algho allows to specify an URL to which the collected information can be sent via the call to a web-service.
In this case, the system uses the field's values as parameters for the call to the service.


\subsection{Computer Vision}
The computer vision functionalities for the described work have been implemented with two open source libraries, OpenCV and TensorFlow. OpenCV~\cite{laganiere2014opencv} provides the state-of-the-art algorithms in this field and, starting from version 4.0, has introduced more advanced features for deep learning. TensorFlow~\cite{abadi2016tensorflow} is a library to develop and train machine learning models, in particular it’s used to create deep neural networks.
Our approach follows a general pipeline composed of three main steps:
\begin{itemize}
\item Dataset creation: several images of the objects of interest are collected and their position is annotated manually by specifying their coordinates;
\item Training the model: a model is trained in order to recognize the objects of our interest and its coordinates within the image. For this purpose, we decided to fine tune the model Faster R-CNN~\cite{ren2015faster} with Inception V2~\cite{szegedy2016rethinking} pre-trained on the COCO dataset~\cite{lin2014microsoft};
\item Using the model: the detection of the requested object through the conversational interface is performed in real time by analysing a video stream received from a video camera.
\end{itemize}

\section{Exprivia's Use Case Scenarios}\label{sec:scenarios}
Exprivia prototyped this general framework in two different use case scenarios, with the main target of enabling communication between all the machines and ICT systems located in a factory in a capillary way, ranging from supply chain systems to administrative ones. 
The ultimate goal is to manage of the entire production life-cycle to a cost saving optimization of each resource that turns into an advantage, not only economical but also competitive, allowing company to play a leading role in the challenge of the future.

{\bf Food Supply Chain}. 
An interesting example of the application of our framework has been made within the food supply chain, in particular referring to the \emph{pasta creation chain}, presented at the DevNet Create 2019 conference in Mountain View (California) in April.
The purpose of the project was to automate a series of activities typical of daily operations, specifically to medium-high difficulty activities that are the cause of most problems in the production life-cycle.
Pasta creation process is very complex and requires a concatenation of different work steps. Many of these are performed manually (e.g. quality control) and typically the machines are not able to communicate with each other: this means that operators and the management cannot have information on the operating status. 
%
Thanks to our framework that includes a chat-bot to communicate with the machinery and computer vision algorithms able to automate the pasta quality control, the communication with management systems enables a two-way exchange of information that automates activities, improving overall operating efficiency.

%


{\bf Coffee Pod Selection with Nuccio}. 
The following solution provides the possibility to use a Fanuc Cobot to select a \emph{coffee pod}. This prototype has been presented at Mobile World Congress 2019, in Barcelona, in February.
The Cobot ``\emph{Nuccio}'' is controlled through the Algho conversational interface.
In particular, the idea was to create a conversational agent focused on a specific knowledge base about coffee.
The resulting bot was able to handle conversations about coffee and about many aspects related to this topic.
Afterwards, a specific ``conversational form'' was developed for collecting a set of information useful for preparing a coffee (taste, aroma, sugar, short or long) and required by the actuator system. 
Finally, the form has been connected with the web-service of the actuator system and linked to the set of intents for which activation was desired.
Thus, the resulting bot was able to handle conversation concerning coffee and if the user request deals with the intent to have a coffee, the linked conversational form allows to collect all the information required by the actuator system to prepare the coffee and to notify via a web service call.
Moreover, Nuccio, through the use of a camera, was able to acquire images and process through the use of Computer Vision algorithms, recognizing exactly the pod to be selected without knowing the position in advance. 
Through the Algho conversational interface, the user is helped and guided in the choice of the most suitable coffee pod, according to his/her tastes.



\section{Conclusion}\label{sec:conclusion}

In line with the main objectives, we contributed to the development and validation of a framework in an operational environment of intelligent robotic systems and HRC. 
In particular, we dealt with conversational interaction technologies useful to perform: (i) high-performance linguistic analysis services based on NLP technologies; (ii) models for the symbiotic human-robot interaction management
; (iii) services and tools for the adaptation of linguistic interfaces with respect to user characteristics. 
The Cobots are close to operating in environments where the presence of man plays a key role. A fundamental characteristic is therefore the Cobot's ability in reacting to textual and vocal commands to properly understand the user's commands. The Cobot's perception is leveraged with  its ability to detect object and understand what there is around him; computer vision processing becomes crucial to the extent of giving Cobots a cognitive profile.
We therefore envision our framework to be fully operable in complex manufacturing systems, in which the collaboration between robot and man is facilitated by advanced AI and cognitive techniques.

We showed how, already today, it is possible to ``humanize'' highly automated processes through a Cobot, collecting and integrating the operational information in the corporate knowledge base. In fact, we believe that in the long term there will be a convergence between automation, AI and IoT, allowing the market to create a full ``Digital Twin'' with an organization that will lead to a strong automation of organizational choices driven by data collected in the field. The digitized organization can then be equipped with its own ``Company Brain'', an AI able to make autonomous complex decisions aimed at maximizing a business goal that, working in a cooperative manner with the company management, will be able to respond much more precisely and quickly to changes in an increasingly unstable and fluid market.



\bibliographystyle{acl}
\bibliography{references}

\begin{thebibliography}{}

\bibitem[\protect\citename{Abadi \bgroup et al.\egroup }2016]{abadi2016tensorflow}
Mart{\'\i}n Abadi, Paul Barham, Jianmin Chen, Zhifeng Chen, Andy Davis, Jeffrey Dean, Matthieu Devin, Sanjay Ghemawat, Geoffrey Irving, Michael Isard, et~al.
\newblock 2016.
\newblock Tensorflow: A system for large-scale machine learning.
\newblock In {\em 12th $\{$USENIX$\}$ Symposium on Operating Systems Design and Implementation ($\{$OSDI$\}$ 16)}, pages 265--283.

\bibitem[\protect\citename{Bongini \bgroup et al.\egroup }2018]{rigutini2018recursive}
Marco Bongini, Leonardo Rigutini, and Edmondo Trentin.
\newblock 2018.
\newblock Recursive neural networks for density estimation over generalized random graphs.
\newblock {\em IEEE Transactions on Neural Networks and Learning Systems}, 29(11):5441--5458.

\bibitem[\protect\citename{Colgate \bgroup et al.\egroup }1996]{colgate1996cobots}
James~Edward Colgate, Michael~A. Peshkin, and Witaya Wannasuphoprasit.
\newblock 1996.
\newblock Cobots: Robots for collaboration with human operators.
\newblock In {\em Proceedings of the International Mechanical Engineering Congress and Exhibition, Atlanta, GA}, volume~58, pages 433--439. Citeseer.

\bibitem[\protect\citename{Djuric \bgroup et al.\egroup }2016]{djuric2016cobot}
Ana~M. Djuric, R.J. Urbanic, and J.L. Rickli.
\newblock 2016.
\newblock A framework for collaborative robot (cobot) integration in advanced manufacturing systems.
\newblock {\em SAE International Journal of Materials and Manufacturing}, 9(2):457--464.

\bibitem[\protect\citename{Dubey and Crowder}2002]{dubey2002finger}
Venketesh~N. Dubey and Richard~M. Crowder.
\newblock 2002.
\newblock A finger mechanism for adaptive end effectors.
\newblock In {\em ASME 2002 International Design Engineering Technical Conferences and Computers and Information in Engineering Conference}, pages 995--1001. American Society of Mechanical Engineers.

\bibitem[\protect\citename{El~Zaatari \bgroup et al.\egroup }2019]{el2019cobot}
Shirine El~Zaatari, Mohamed Marei, Weidong Li, and Zahid Usman.
\newblock 2019.
\newblock Cobot programming for collaborative industrial tasks: An overview.
\newblock {\em Robotics and Autonomous Systems}, 116:162--180.

\bibitem[\protect\citename{Ionescua and Schlunda}2019]{ionescua2019participatory}
Tudor~B. Ionescua and Sebastian Schlunda.
\newblock 2019.
\newblock A participatory programming model for democratizing cobot technology in public and industrial fablabs.
\newblock {\em Procedia CIRP}, 81:93--98.

\bibitem[\protect\citename{Ivorra \bgroup et al.\egroup }2018]{ivorra2018multimodal}
Eugenio Ivorra, Mario Ortega, Mariano Alca{\~n}iz, and Nicol{\'a}s Garcia-Aracil.
\newblock 2018.
\newblock Multimodal computer vision framework for human assistive robotics.
\newblock In {\em 2018 Workshop on Metrology for Industry 4.0 and IoT}, pages 1--5. IEEE.

\bibitem[\protect\citename{Lagani{\`e}re}2014]{laganiere2014opencv}
Robert Lagani{\`e}re.
\newblock 2014.
\newblock {\em OpenCV Computer Vision Application Programming Cookbook Second Edition}.
\newblock Packt Publishing Ltd.

\bibitem[\protect\citename{Lin \bgroup et al.\egroup }2014]{lin2014microsoft}
Tsung-Yi Lin, Michael Maire, Serge Belongie, James Hays, Pietro Perona, Deva Ramanan, Piotr Doll{\'a}r, and C.~Lawrence Zitnick.
\newblock 2014.
\newblock Microsoft {COCO}: Common objects in context.
\newblock In {\em European conference on computer vision}, pages 740--755. Springer.

\bibitem[\protect\citename{Melacci \bgroup et al.\egroup }2018]{rigutini2018enhancingwsd}
Stefano Melacci, Achille Globo, and Leonardo Rigutini.
\newblock 2018.
\newblock Enhancing modern supervised word sense disambiguation models by semantic lexical resources.
\newblock In {\em Proceedings of the Eleventh International Conference on Language Resources and Evaluation ({LREC}-2018)}, Miyazaki, Japan, May. European Languages Resources Association (ELRA).

\bibitem[\protect\citename{Ren \bgroup et al.\egroup }2015]{ren2015faster}
Shaoqing Ren, Kaiming He, Ross Girshick, and Jian Sun.
\newblock 2015.
\newblock Faster{R-CNN}: Towards real-time object detection with region proposal networks.
\newblock In {\em Advances in neural information processing systems}, pages 91--99.

\bibitem[\protect\citename{Szegedy \bgroup et al.\egroup }2016]{szegedy2016rethinking}
Christian Szegedy, Vincent Vanhoucke, Sergey Ioffe, Jon Shlens, and Zbigniew Wojna.
\newblock 2016.
\newblock Rethinking the inception architecture for computer vision.
\newblock In {\em Proceedings of the IEEE conference on computer vision and pattern recognition}, pages 2818--2826.

\bibitem[\protect\citename{Wenger}2014]{wenger2014artificial}
Etienne Wenger.
\newblock 2014.
\newblock {\em Artificial intelligence and tutoring systems: computational and cognitive approaches to the communication of knowledge}.
\newblock Morgan Kaufmann.

\bibitem[\protect\citename{Zue and Glass}2000]{zue2000conversational}
Victor~W Zue and James~R Glass.
\newblock 2000.
\newblock Conversational interfaces: Advances and challenges.
\newblock {\em Proceedings of the IEEE}, 88(8):1166--1180.

\end{thebibliography}

\end{document}